\documentclass[letterpaper, 10 pt, conference]{ieeeconf}  
\IEEEoverridecommandlockouts                              
\usepackage{graphics}    
\usepackage{times}       
\usepackage{amsmath}     
\usepackage{amssymb}     
\usepackage{graphicx}
\usepackage{algorithm}
\usepackage[noend]{algpseudocode}
\usepackage{subfigure}
\usepackage{gensymb}
\usepackage{booktabs}
\usepackage{caption}

\usepackage[hyphens]{url}
\usepackage{hyperref}

\usepackage[font=small]{caption}

\def\secref#1{Sec.~\ref{#1}}
\def\figref#1{Fig.~\ref{#1}}
\def\tabref#1{Tab.~\ref{#1}}
\def\eqref#1{Eq.~(\ref{#1})}
\def\algref#1{Alg.~\ref{#1}}
\def\lineref#1{line~\ref{#1}}

\newcommand\etal{\emph{et al.}}
\newcommand\circled[1]{\textcircled{\footnotesize #1}}

\title{\LARGE \bf Visual Servoing-based Navigation for Monitoring Row-Crop Fields}

\author{Alireza Ahmadi \and Lorenzo Nardi \and Nived Chebrolu \and Cyrill Stachniss
  \thanks{All authors are with the University of Bonn, Germany. This work has partly been supported by the German Research Foundation under Germany's Excellence Strategy, EXC-2070 - 390732324 (PhenoRob).
  }
}

\begin{document}
\maketitle
\thispagestyle{empty}
\pagestyle{empty}

\begin{abstract}

Autonomous navigation is a pre-requisite for field robots to carry out precision agriculture tasks. Typically, a robot has to navigate through a whole crop field several times during a season for monitoring the plants, for applying agro-chemicals, or for performing targeted intervention actions. In this paper, we propose a framework tailored for navigation in row-crop fields by exploiting the regular crop-row structure present in the fields. Our approach uses only the images from on-board cameras without the need for performing explicit localization or maintaining a map of the field and thus can operate without expensive RTK-GPS solutions often used in agriculture automation systems. Our navigation approach allows the robot to follow the crop-rows accurately and handles the switch to the next row seamlessly within the same framework. We implemented our approach using C++ and ROS and thoroughly tested it in several simulated environments with different shapes and sizes of field. We also demonstrated the system running at frame-rate on an actual robot operating on a test row-crop field. The code and data have been published.

\end{abstract}

\section{Introduction}
\label{sec:intro}

\begin{figure}[t]
  \centering
  \includegraphics[width=0.98\linewidth]{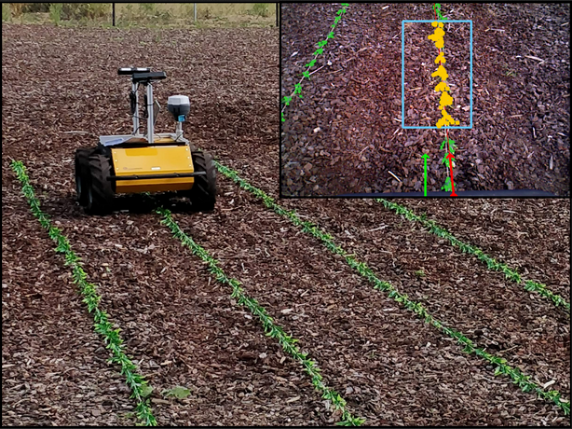}
  \caption{Robot navigation in a test row-crop field. 
           Top-right: on-board camera image. The visual servoing based controller executes the velocity control that brings the crop row~(red arrow) to the center of the camera image~(green arrow). The blue box shows the sliding window used
           for tracking the row along which the robot navigates.}
  \label{fig:motivation}
\end{figure}

Autonomous agricultural robots have the potential to improve  farm
productivity and to perform targeted field management activities. In crop
fields, agricultural robots are typically used to perform monitoring
tasks~\cite{kusumam2017jfr},~\cite{nakarmi2014biosyseng} or targeted
intervention such as weed control~\cite{wu2019icra},~\cite{mccool2018ral}.
Several crops such as maize, sugar beet, sunflower, potato, soybean,
and many others are arranged along multiple parallel rows in the fields as
illustrated in~\figref{fig:motivation}. This arrangement facilitates
cultivation, weeding, and other farming operations. For accomplishing such
tasks,  robots must be able to autonomously navigate through the crop-rows
repeatedly in the field.

Currently, a popular solution for navigating autonomously in fields is to use a
high-precision, dual-frequency RTK-GNSS receiver to guide the robot along
pre-programmed paths. However, the high cost of these systems and vulnerability
to outages has led to an interest in solutions using observations from on-board
sensors. Such solutions typically use observations from a laser scanner or a
camera to localize the robot in the environment and then navigate along crop
rows, often with the help of a map. The crop field scenario poses serious
challenges to such systems due to high visual aliasing in the fields and
lack of reliable sensor measurements of identifiable landmarks to support
localization and mapping tasks. Additionally, as the field is constantly
changing due to the growth of plants, a map of the field needs to be updated
several times during a crop season.

In this paper, we address navigation in row-crop fields 
only based on camera observations and by exploiting the row structure
inherent in the field to guide the robot and cover the field. An example
illustrating our robot navigating along a crop-row is shown in \figref{fig:motivation}. 
We aim at controlling the robot without explicitly
maintaining a map of the environment or performing localization in a global
reference frame.

The main contribution of this paper is a novel navigation system for
agricultural robots operating in row-crop fields. We present a visual servoing-based 
controller that controls the robot using local directional features
extracted from the camera images. This information is obtained from the crop-row
structure, which is continuously tracked through a sliding window.  Our approach
integrates a switching mechanism to transition from one row to the next one when
the robot reaches the end of a crop-row. By using a pair of cameras, the robot
is able to enter a new crop-row within a limited space and avoids making sharp
turns or other complex maneuvers.
As our experiments show, the proposed approach allows a robot to (i)~autonomously navigate through
row-crop fields without the maintaining any global reference maps, (ii)~monitor the crops in the fields with a high coverage by accurately following
the crop-rows, (iii)~is robust to fields with different row structures and
characteristics, as well as to critical user-defined parameters.

Note that the source code of our approach, the data from the real-world experiments as
well as the simulated environment are available at: 
\url{http://github.com/PRBonn/visual-crop-row-navigation}.

\begin{figure*}[t]
  \centering
  \includegraphics[width=0.92\linewidth]{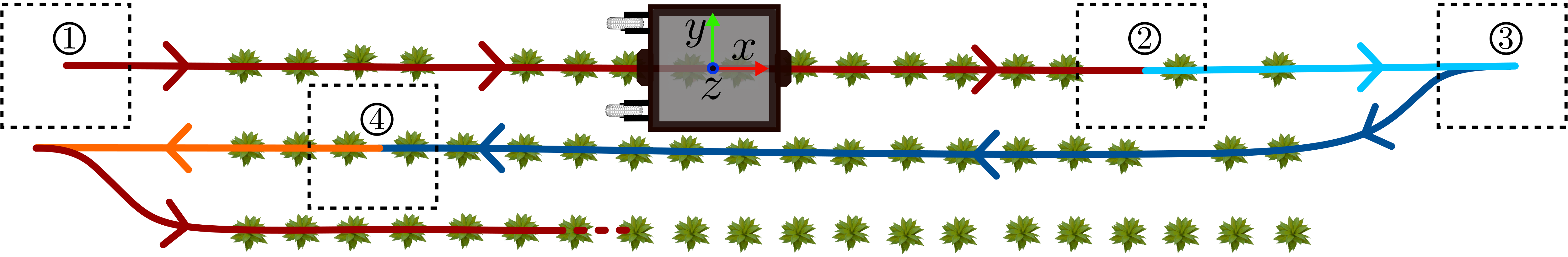}
  \caption{Scheme for navigation in a crop field: the robot 
  enters the field and navigates along a crop row~(\circled{1}), exits the row~(\circled{2}), transitions to the next crop row~(\circled{3}), and exits the row on the opposite side~(\circled{4}).}
\label{fig:task}
\end{figure*} 

\section{Related Work}
\label{sec:related}

Early autonomous systems for navigation in crop fields such as the one
developed by Bell \cite{bell2000cea} or Thuilot~\etal~\cite{thuilot2001iros}
are based on GNSS while others use visual fiducial markers
\cite{olson2011icra-aara} or artificial beacons \cite{leonard1991tra}. More
recently, agricultural robots equipped with a suite of sensors including GNSS
receiver, laser-scanner, and camera have been used for precision agriculture
tasks such as selective spraying ~\cite{underwood2015icra} or targeted
mechanical intervention~\cite{imperoli2018ral}. Dong~\etal~\cite{dong2017icra}
as well as Chebrolu~\etal~\cite{chebrolu2019icra} address the issue of
changing appearance of the crop fields and proposed localization systems,
which fuse information from several on-board sensors and prior aerial maps to
localize over longer periods of time. While most of these systems allow for
navigation in crop fields accurately, they require either additional
infrastructure or reference maps for navigation. In contrast to that, our
approach only requires local observations of the crop rows from obtained from
a camera.

Other approaches also exploit the crop row structure in the fields and
developed vision based guidance systems for controlling the
robot~\cite{billingsley1997cea} and perform weeding operations in between the
crop rows~\cite{aastrand2005mec}. These methods require a reliable crop row
detection system as they use this information for computing the control signal
for the robot. As a result, several works focus on detecting crop rows from
images under challenging conditions.
Winterhalter~\etal~\cite{winterhalter2018ral} propose a method for detecting
crop rows based on hough transform and obtain robust detections even at an
early growth stage when are plants are very small.
English~\etal~\cite{english2014icra} and
S{\o}gaard~\etal~\cite{sogaard2003cae} show reliable crop row detections in
the presence of many weeds and under challenging illumination conditions.
Other works such as~\cite{midtiby2012bs}, \cite{haug2014ias},
\cite{kraemer2017iros} aim  at estimating the stem locations of the plants
accurately. While we use a fairly simple method for crop row detection in this
paper, more robust methods for detection can be easily integrated in our
approach for dealing with challenging field conditions with a high percentage
of weeds. The proposed controller is independent of the used detection
technique.

Traditionally, visual servoing techniques~\cite{espiau1992tra} are used for
controlling robotic arms and manipulators.  These techniques aim to control
the motion of the robot by directly using vision data in the control loop.
Cherubini~\etal~\cite{cherubini2008iccarv},~\cite{cherubini2008iros},~\cite{ma1999tra}
propose visual servoing techniques for the controlling mobile robots along
continuous paths. De Lima~\etal~\cite{delima2014itsc} apply these visual
servoing techniques to autonomous cars for following lanes in urban scenarios,
whereas Avanzini~\etal~\cite{avanzini2010iros} control a platoon of cars using
a guiding vehicle. We have built upon these ideas to develop our controller
for crop field navigation including a mechanism to transition from one row to
the next row within the same framework.

\section{Navigation Along Row-Crop Fields}

\begin{figure*}[th]
\centering
\subfigure[Our self-built robot]{
    \includegraphics[height=3.33cm]{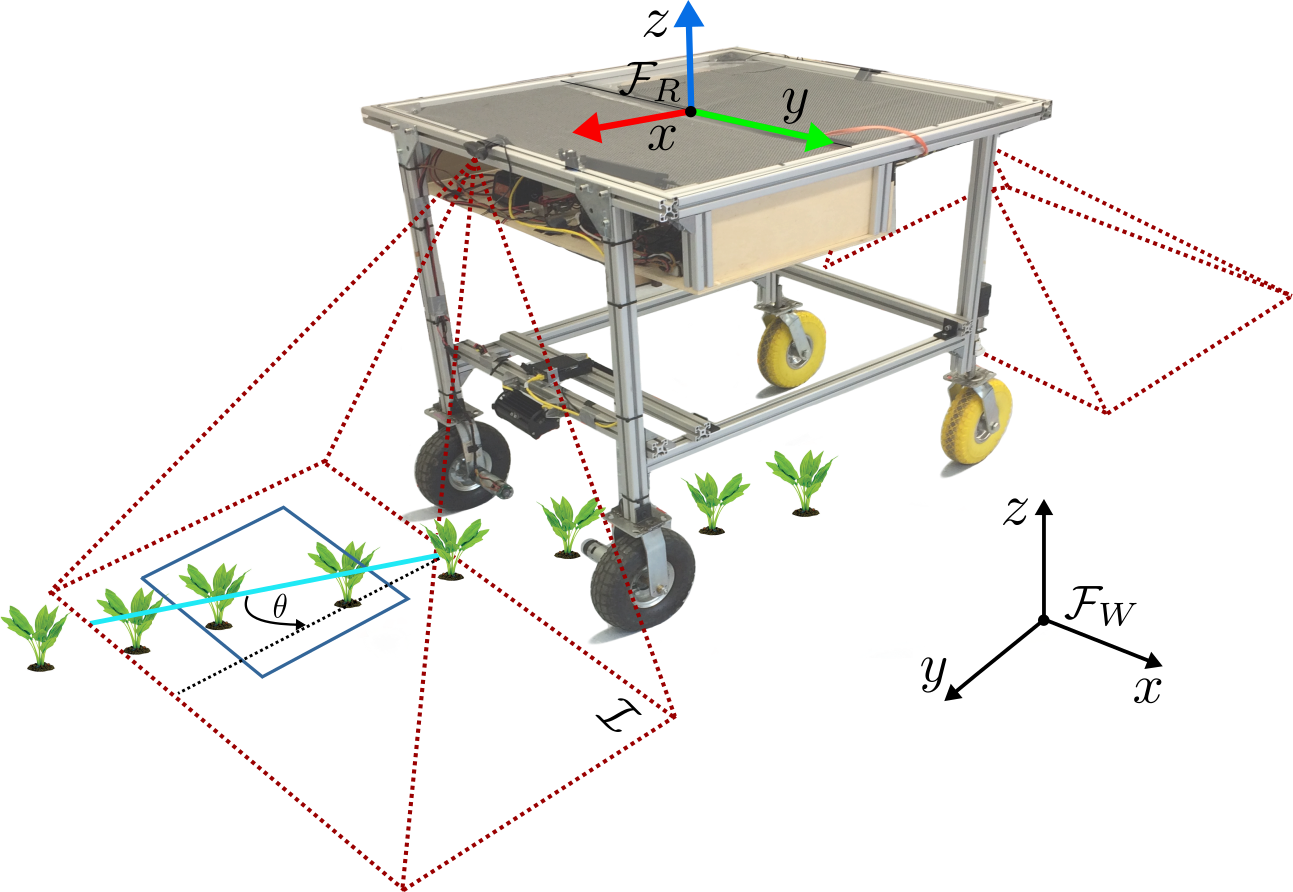}\label{fig:robot:3d}}\hspace{5px}
\subfigure[Robot side view]{
    \includegraphics[height=3.33cm]{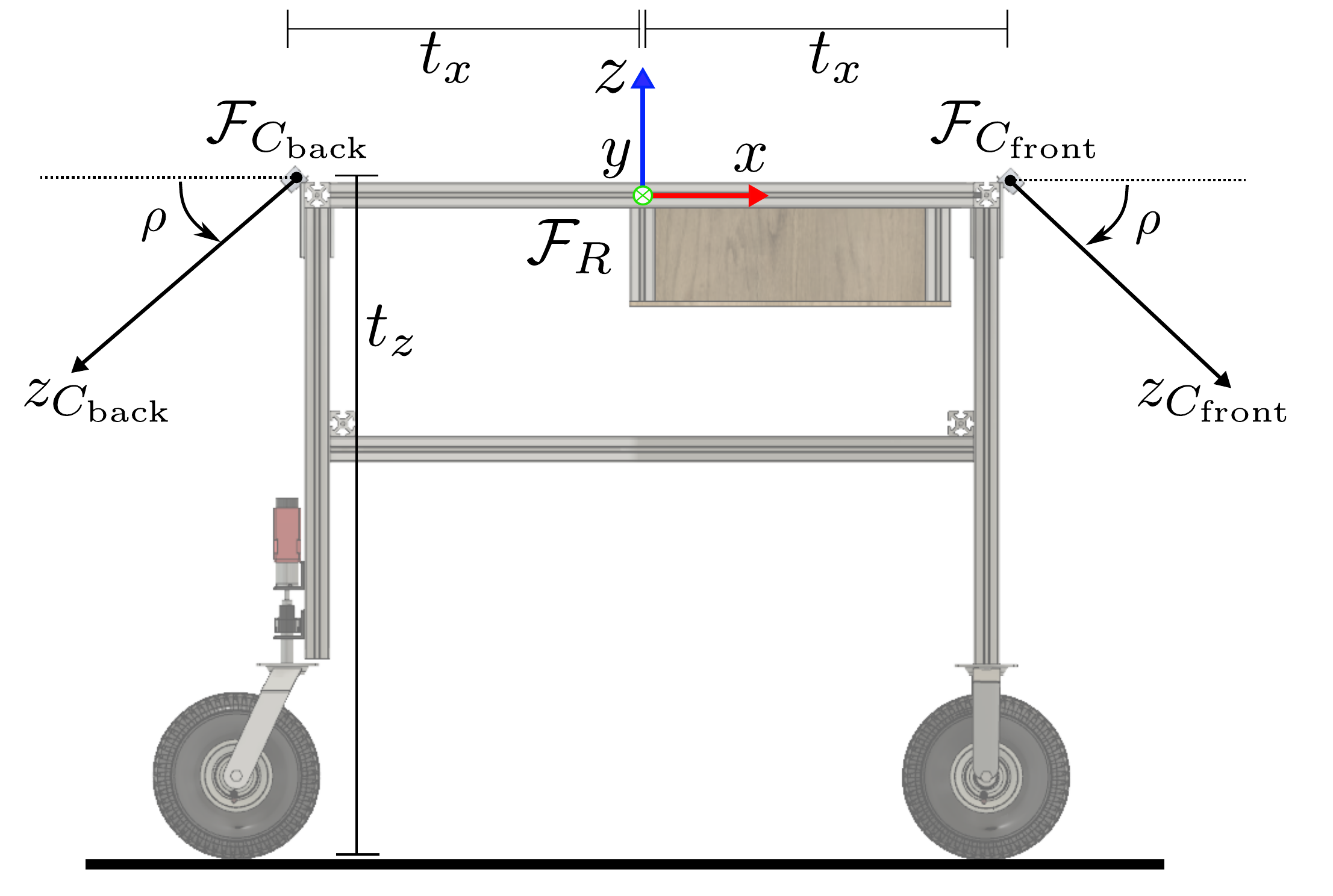}\label{fig:robot:side}}
\subfigure[Camera image]{
    \includegraphics[height=3.33cm]{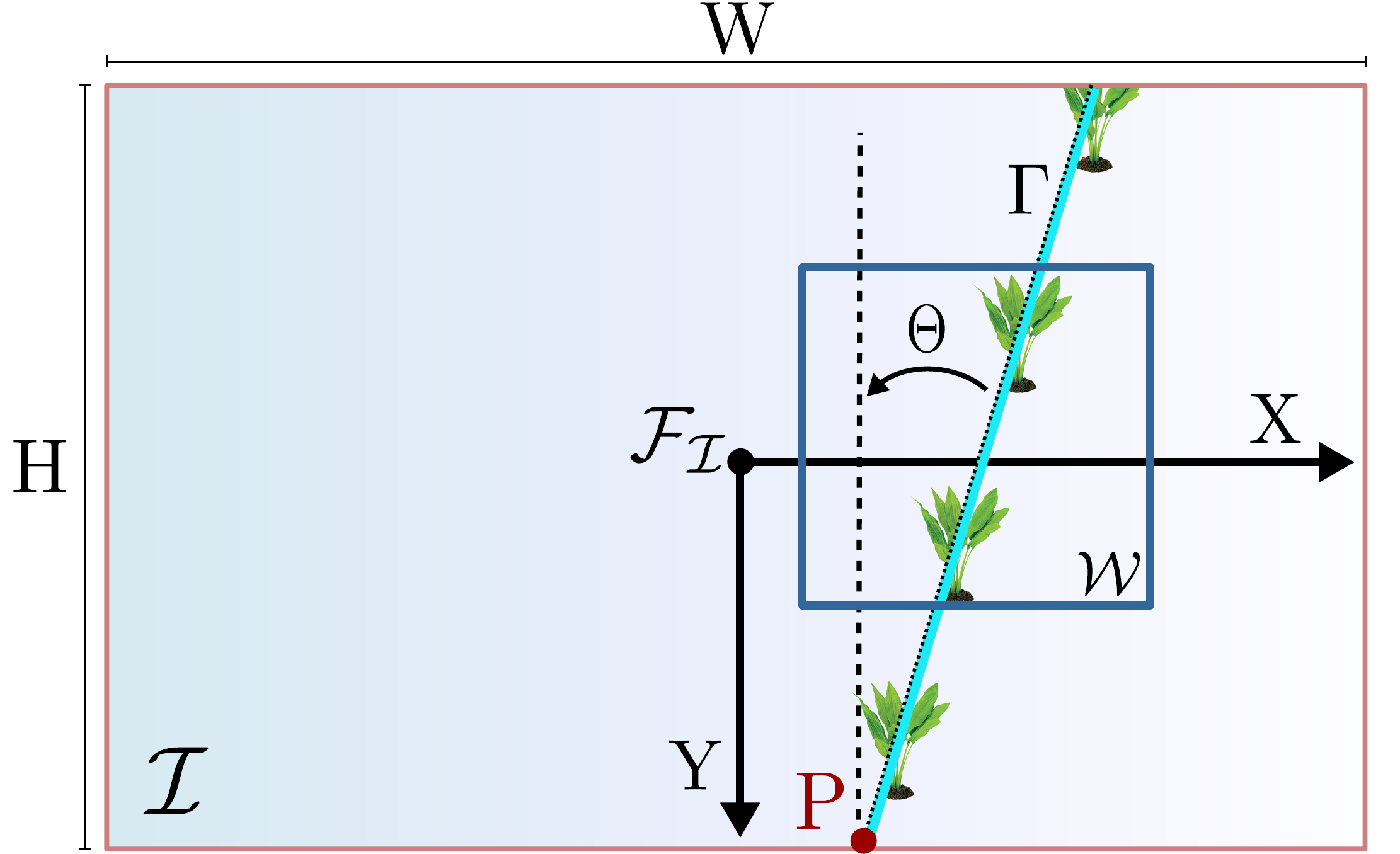}\label{fig:robot:image}}
\caption{
Robot, frames and variables. The robot navigates by following the path~(cyan) along the crop row.
In \protect\subref{fig:robot:3d}, $\mathcal{F}_{W}$ and $\mathcal{F}_{R}$ are world and robot frames, $\theta$ is the orientation of the robot in $\mathcal{F}_{W}$.  
In \protect\subref{fig:robot:side}, $\mathcal{F}_{C_\mathrm{front}}$, $\mathcal{F}_{C_\mathrm{back}}$ are the front and back camera frames. The cameras are mounted at an offset~$t_x$ from the robot center~$\mathcal{F}_{R}$ and~$t_z$ above the ground, and with tilt~$\rho$.  
In \protect\subref{fig:robot:image}, $\mathcal{F_I}$ is the image frame and $s = [X,\,Y,\,\Theta]$
 is the image features computed from the crop row.
}
\label{fig:robot}
\end{figure*} 

In this paper, we consider a mobile robot that navigates in row-crop fields to
perform tasks such as monitoring of the crops or removing the weeds and has to
cover the field row by row. Thus, the robot must be able to navigate
autonomously along all the crop rows in the field.

\subsection{Navigation Along Crop Rows} 

In row-crop fields, crops are arranged along multiple parallel curves. We take
advantage of such an arrangement to enable a mobile robot to autonomously
navigate and monitor the crops in the field. The main steps to achieve this are
illustrated in~\figref{fig:task}. The robot starts from one of the corners of
the field~(\circled{1}), enters the first crop row and follows it until the
end~(\circled{2}). As it reaches the end of a row, it needs to enter the next
one and follow it in the opposite direction~(\circled{3}). This sequence of
behaviors is repeated to navigate along all the crop rows in the field. We
achieve this by using a visual-based navigation system that integrates all of
these behaviors which leads the robot to autonomously navigate and monitor the
crops in a row-crop field.


\subsection{Robotic Platform}

We consider a mobile robotic platform that is able to navigate on crop fields by
driving seamlessly forward and backwards. We equip this robot with two
cameras~$\mathcal{F}_{C_\mathrm{front}}$ and~$\mathcal{F}_{C_\mathrm{back}}$
mounted respectively looking to the front and to the back of the robot as
illustrated in~\figref{fig:robot}. The cameras are symmetric with respect to the
center of rotation of the robot denoted as~$\mathcal{F}_{R}$. The cameras have a
fixed tilt angle~$\rho$ and are positioned on the robot at a height~$t_z$ from
the ground and with an horizontal offset~$t_y$ from~ $\mathcal{F}_{R}$. A camera
image is illustrated in~\figref{fig:robot:image}, where~$W$ and~$H$ are
respectively its width and height in pixels.

\section{Our Navigation Approach}
\label{sec:main}


We propose to use a visual-based navigation system that relies only local visual
features and exploits the arrangement of the crops in fields to autonomously
navigate in row-crop fields without requiring an explicit map. Our visual-based
navigation system builds upon the image-based visual servoing controller by
Cherubini~\etal~\cite{cherubini2008iccarv} and extends it by considering an
image that presents multiple crop-rows and integrating a mechanism for switching
to the next row. 

\subsection{Visual Servoing Controller}
\label{sec:main:vsc}

Visual servoing allows for controlling a robot by processing visual
information. Cherubini~\etal~\cite{cherubini2008iccarv} propose an
image-based visual servoing scheme that allows a mobile robot equipped with a
fixed pinhole camera to follow a continuous path on the ground. It uses a
combination of two primitive image-based controllers to drive the robot to the
desired configuration.

We define the robot configuration as~$q = [x,\,y,\,\theta]^{T}$. The control
variables are the linear and the angular velocity of the robot~$u =
[v,\,\omega]^{T}$. We impose that the robot moves with a fixed constant
translational velocity~$v = v^{*}$. Thus, our controller controls only the
angular velocity~$\omega$.

The controller computes the controls~$u$ by minimizing the error~$e = s -
s^{*}$, where~$s$ is a vector of features computed on the camera image
and~$s^{*}$ is the desired value of the corresponding features.The state
dynamics are given by:
\begin{eqnarray}
  \dot{s} = J\,u = J_v\,v^{*} + J_\omega\,\omega,
\end{eqnarray}
where~$J_v$ and $J_\omega$ are the columns of the Jacobian~$J$ that relate~$u$ to~$\dot{s}$, 
the controller computes the controls by applying the feedback control:
\begin{eqnarray}
  \omega = -J_\omega^{+}\,(\lambda e + J_v v^{*}), \quad \lambda > 0,
  \label{eq:controllaw}
\end{eqnarray}
where~$J_\omega^{+}$ indicates the Moore-Penrose pseudo-inverse of~$J_\omega$.

From the camera image, we compute an image feature~$s = [X,\,Y,\,\Theta]$
illustrated in~\figref{fig:robot:image} where~$P = [X,\,Y]$ is the position of
the first point along the visible path and~$\Theta$ is the orientation of the
tangent~$\Gamma$ to the path. We use uppercase variables to denote the 
quantities in the image frame~$\mathcal{I}$. The controller computes a
control~$u$ such that it brings~$P$ to the bottom center of the image~$s^{*} =
[0,\,\frac{H}{2},\,0]$.  The desired configuration corresponds to driving the
robot along the center of the path. The image feature and its desired position
are illustrated in~\figref{fig:motivation}~(top-right).

The interaction matrix~$L_s$ allows for relating the dynamics of the image
features~$s$ to the robot velocity in the camera frame~$u_c$. The velocity in
the camera frame~$u_c$ can be expressed as a function of the robot
velocity~$u$ as~$u_c =\,^CT_R\,u$, where~$^CT_R$ is the homogeneous
transformation from~$\mathcal{F}_R$ to~$\mathcal{F}_C$. Therefore, we can
write the relation between the image feature dynamics $\dot{s}$ and the robot
controls $u$ as:
 \begin{eqnarray}
  \dot{s} = L_s\,u_c = L_s\,^CT_R\,u.
\end{eqnarray}


\subsection{Crop Row Detection for Visual Servoing}
\label{sec:main:vsrow}

The visual-servoing approach described in the previous section allows the
robot to follow a continuous path drawn on the ground. In fields, we can
exploit the arrangement of the crops in rows to enable the robot to navigate
using a similar visual-servoing scheme.

To navigate along a crop-row, we extract the curve along which the crops are
arranged. To this end, for each new camera image we first compute the
vegetation mask using the Excess Green Index (ExG)~\cite{woebbecke1995asae}
often used in agricultural applications. It is given by $I_{ExG}= 2I_G - I_R
-I_B$ where $I_R$, $I_G$ and $I_B$ correspond to the red, green and blue
channels of the image. For each connected component in the vegetation mask, we
compute a center point of the crop. We then estimate the path curve along
which the robot should navigate by computing the line that best fits all the
center points using a robust least-squares fitting method. This procedure
allows for continuously computing a path curve in the image that the robot can
follow using the visual-servoing controller described in
~\secref{sec:main:vsc}. In this paper, we use a fairly straight-forward
approach to detect the crop rows as our main focus has been on the design of
the  visual servoing controller and more sophisticated detection methods are
easily implementable. 

Typically, fields are composed by number of parallel crop-rows. Thus, multiple
rows can be visible at the same time in the camera image. This introduces
ambiguity to identify the curve that the robot should follow. This ambiguity may
cause the robot to follow a different crop-row before reaching the end of the
current one. If this is case, there is no guarantee that the robot will navigate
through the whole field. To remove this ambiguity, we use a sliding
window~$\mathcal{W}$ of fixed size in the image that captures the row that the
robot is following, as illustrated on the bottom of~\figref{fig:motivation}. For
every new camera image we update the position of the window~$\mathcal{W}$ by
centering it at the average position of the crops detected in that frame.
Updating this window continuously allows for tracking a crop row and ensures
that the robot follows it up to its end.

\subsection{Scheme for Autonomous Navigation in Crop-Row Fields}
\label{sec:main:rownav}


The visual-based navigation system described in the previous section allows the
robot to navigate by following a single  crop row. To cover the whole field, the
robot should be able to transition to the next row upon reaching the end of the
current one. However, as the rows are not connected to each other, the robot has
no continuous path curve to follow over the whole field. Therefore, we introduce
a visual-based navigation scheme that allows the robot to follow a crop-row, to
exit from it, and to enter the next one by exploiting both the cameras mounted
on the robot and its ability to drive both in forward and backward directions.
Our scheme to navigate in crop fields is illustrated in~\algref{alg:scheme}.

The visual-servoing controller described in section~\secref{sec:main:vsc} uses
the image of one camera to compute the controls.  We extend this approach to
using both the front camera~$\mathcal{F}_{C_\mathrm{front}}$ and the back
camera~$\mathcal{F}_{C_\mathrm{back}}$. We set in turn the camera used by the
visual-servoing controller, which we refer to as the primary
camera~$\mathrm{cam}_\mathcal{P}$. Whereas, we denote the other camera as the
secondary camera~$\mathrm{cam}_\mathcal{S}$. We define the size of the sliding
window~$\mathcal{W}$ used by the controller and a shift offset to capture the
next crop row based on the tilt angle of the camera~$\rho$ and an estimate of
the distance between the crop rows~$\delta$.

\begin{algorithm}[t]
\caption{Crop row navigation scheme }
\begin{algorithmic}[1]
\State $\mathcal{W}\gets\Call{initializeWindow}{}$ \Comment{Initialization.} \label{alg:scheme:init}
\Repeat \Comment{Control loop.} \label{alg:scheme:loop}
  \State $\mathrm{crops_\mathcal{P}} \,\gets \Call{detectCrops}{\mathrm{cam_\mathcal{P}}}$ \label{alg:scheme:detect}
  \State $\mathrm{crops_\mathcal{W}} \gets \Call{cropsInWindow}{\mathrm{crops_\mathcal{P}},\,\mathcal{W}}$ \label{alg:scheme:select}
  \If{$\Call{isEmpty}{\mathrm{crops_\mathcal{W}}}$}
    \If{$\Call{isEmpty}{\textsc{detectCrops}(\mathrm{cam_\mathcal{S}})}$} \label{alg:scheme:next}
        \State $\mathcal{W}\gets\Call{shiftWindow}{}$ \Comment{Enter next row.}
      \Else \Comment{Exit row.} \label{alg:scheme:exit}
        \State $\Call{switchCameras}{\mathrm{cam_\mathcal{P}},\mathrm{cam_\mathcal{S}}}$ \label{alg:scheme:switch}
        \State $\mathcal{W}\gets\Call{initializeWindow}{}$
        \State $\mathrm{crops_\mathcal{P}} \,\gets \Call{detectCrops}{\mathrm{cam_\mathcal{P}}}$
      \EndIf
        \State $\mathrm{crops_\mathcal{W}} \gets \Call{cropsInWindow}{\mathrm{crops_\mathcal{P}},\,\mathcal{W}}$
  \EndIf
  \State $\Call{followCropRow}{\mathrm{crops_\mathcal{W}}}$ \label{alg:scheme:navigate}
  \State $\mathcal{W}\gets\Call{updateWindow}{\mathrm{crops_\mathcal{W}}}$ \label{alg:scheme:updatew}
\Until{$\Call{isEmpty}{\mathrm{crops_\mathcal{W}}}$} \Comment{Stop navigation.}
\end{algorithmic}
\label{alg:scheme}
\end{algorithm}

Our navigation scheme assumes that the starting position of the robot is at
one of the corners of the field (see for example \circled{1}
in~\figref{fig:task}). We initially set the camera looking in the direction of
the field as the primary camera~$\mathrm{cam}_\mathcal{P}$ and initialize the
position of the window~$\mathcal{W}$ at the center of the image
(\lineref{alg:scheme:init} of~\algref{alg:scheme}).

In the control loop (\lineref{alg:scheme:loop}), we first detect the centers
of the crops~$\mathrm{crops}_\mathcal{P}$ in the image of the primary
camera~(\lineref{alg:scheme:detect}) using the approach described
in~\secref{sec:main:vsrow}. We select the crops in the image that lie within
the window~$\mathcal{W}$,
$\mathrm{crops}_\mathcal{W}$~(\lineref{alg:scheme:select}). The robot
navigates along the crop row by computing the line that fits
the~$\mathrm{crops}_\mathcal{W}$ and follows it using the visual servoing
controller (\lineref{alg:scheme:navigate}). Then, it updates the position of
the sliding window~$\mathcal{W}$ in the image at the average position of
the~$\mathrm{crops}_\mathcal{W}$ (\lineref{alg:scheme:updatew}). This
corresponds to the robot following the red path in~\figref{fig:task}. When the
robot approaches the end of the row (\circled{2}), the primary camera does not
see crops anymore as it is tilted to look forward. 

In this position, the secondary camera can still see the crops belonging to
current row (\lineref{alg:scheme:exit}). Therefore, we switch the primary and
secondary camera, re-initialize the window, re-compute the detected crops, and
drive the robot in the opposite direction to which the primary camera is
facing. This setup guides the robot to exit the crop row (light blue path
in~\figref{fig:task}) until it does not detect crops anymore in the
window~$\mathcal{W}$~(\circled{3}). At this point, the secondary camera also
does not see any crops and the robot needs to enter in the next crop row.
Therefore, we shift the sliding window in the direction of the next row
(\lineref{alg:scheme:next}) to capture the crops in it. By doing this, the
robot starts tracking the next crop row and can navigate by following it (blue
path in~\figref{fig:task}). When no crops are present in the sliding
window~(\circled{4}), the robot switches the camera as in~\circled{2}, exits
the row and shift~$\mathcal{W}$ to start following the next one. This control
loop repeats until the robot reaches the end of the field and can not see any
crop row with both its cameras.

Note that our navigation scheme allows the robot to transition from one crop row
to the next one only by switching the cameras and without requiring the robot to
perform a complex maneuver to enter the next row. Furthermore, following our
navigation scheme the robot requires a smaller space for maneuvering than the
one that it would require to perform a sharp U-turn.

\section{Experimental Evaluation}
\label{sec:exp}

The experiments are designed to show the capabilities of our method for navigation in row-crop fields and to support our key claims, which are:
(i)~autonomous navigation through row-crop fields without the need of maintaining any global reference map or 
an external positioning system such as GNSS,
(ii)~monitoring the crops with a high coverage by accurately following the rows in fields with different row structures,
(iii)~robustness to fields with varying properties and to the input parameters provided by the user.

\subsection{Experimental Setup}

In the experiments, we consider our self-built agricultural robot as well as a
Clearpath Husky. Both robots are equipped with  two monocular cameras placed in
the front and the back of the robot with the camera tilt $\rho$ set
to~75\degree.  Our agricultural robot uses a laptop as the main computational
device along with a Raspberry~Pi~3 as a device communication manager. We
implemented our approach on a real robot using~C++ and~ROS. We also created a
simulated version of the robot in Gazebo, which is built on a 1:1 scale and has
the same kinematics as the real robot. We have generated several simulated crop
fields of different shapes and sizes and evaluated our navigation system both on
the simulated crop fields, as well as on the real robot. The experiments with
the Clearpath Husky are provided here as it is a common platform in the robotics
community and thus easier for the  reader to interpret the results. 

\subsection{Navigation in Crop Fields}

\begin{figure}[t]
\centering
  \includegraphics[width=\linewidth]{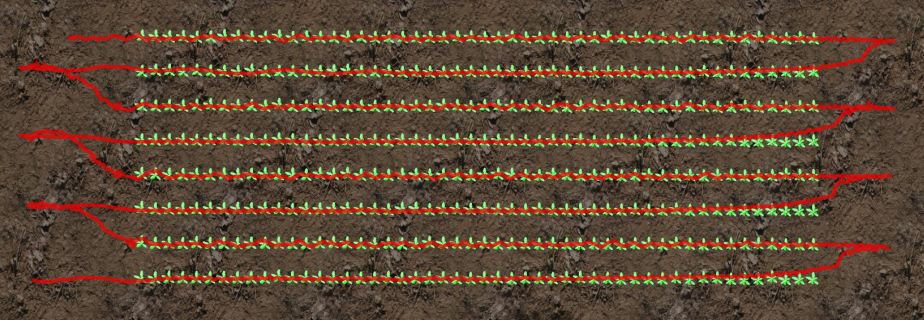}
  \caption{Trajectory of the robot following our visual-based
  navigation scheme in a simulated field environment.}
  \label{fig:exp1:traj}
  \vspace{1em}
  \includegraphics[width=\linewidth]{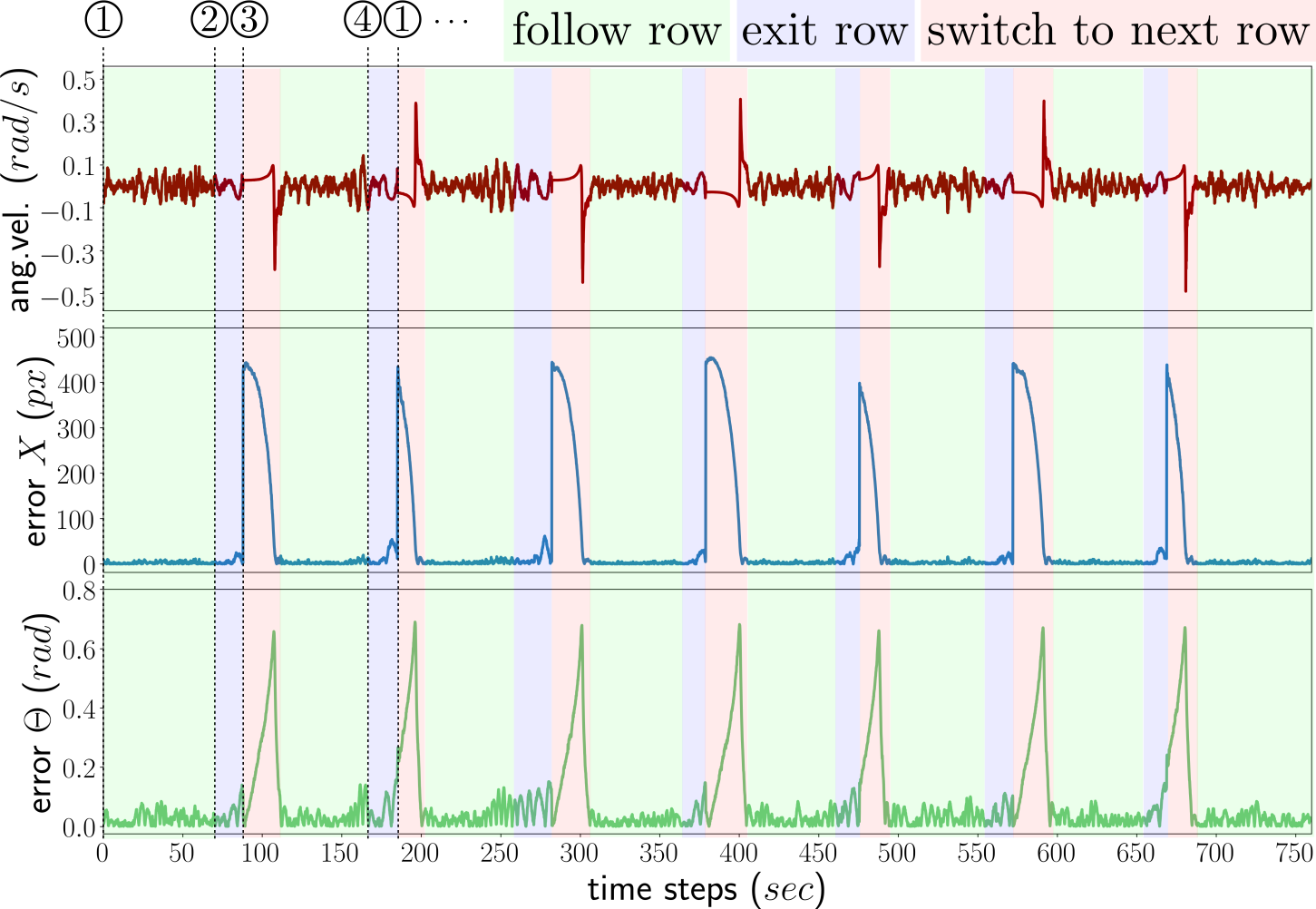}
  \caption{Angular velocity control~(top), error in~$X$~(middle) and error in~$\Theta$~(bottom) computed by the visual servoing controller to navigate along the trajectory illustrated in~\figref{fig:exp1:traj}. We highlighted the steps of our navigation scheme as in~\figref{fig:task} and the robot behaviors.}
  \label{fig:exp1:err}
\end{figure}

The first experiment is designed to show that we are able to autonomously
navigate through crop fields using our navigation system. To this end, we use
our simulated robot that navigates in a crop field environment in Gazebo. We
consider a test field with a dimension of~20\,m$\times$10\,m composed by
8~rows as illustrated in~\figref{fig:exp1:traj}. The rows have an average
crop-row distance of~50\,cm and a standard deviation of~5\,cm. The crops were
distributed along the row with a gap ranging from~5\,cm to~15\,cm  to mimic real
crop-rows. In our setup, the robot starts at the beginning of the first crop row
in top left corner of the field. The goal consists of reaching the opposite
corner by navigating through each row. 

The trajectory along which the robot navigated is illustrated
in~\figref{fig:exp1:traj}. The robot was successfully able to follow each of the
rows until the end and transition to the next ones until it covered the whole
field. At the end of the row, the robot was able to transition to the next row within
an average maneuvering space of around 1.3\,m. Thus, our navigation scheme
allows the robot to enter the next row using a limited maneuvering space which
is often a critical requirement while navigating in a field. 

In~\figref{fig:exp1:err}, we illustrate  the error signals in~$X$
and~$\Theta$, which was used by the visual servoing controller to compute the
angular velocity~$\omega$ for the robot. Both, the error signals in~$X$
and~$\Theta$ show peaks at the times which correspond to the transition to the
next crop row (see for example~\circled{3}). This is the normal behavior
whenever a new row is selected and the robot must align itself to the new row.
The controller compensates for this error using large values of~$\omega$ as
shown in~\figref{fig:exp1:err}~(top). Also, note that the direction
of~$\omega$ is flipped at the end of each row since the robot alternates
between a left and right turn to go through the crop rows. In this experiment,
we demonstrated that our navigation scheme allows the robot to autonomously
monitor a row-crop field by accurately following the crop-rows.

\subsection{Field Coverage Analysis}

\begin{figure}[t]
\centering
\includegraphics[width=0.92\linewidth]{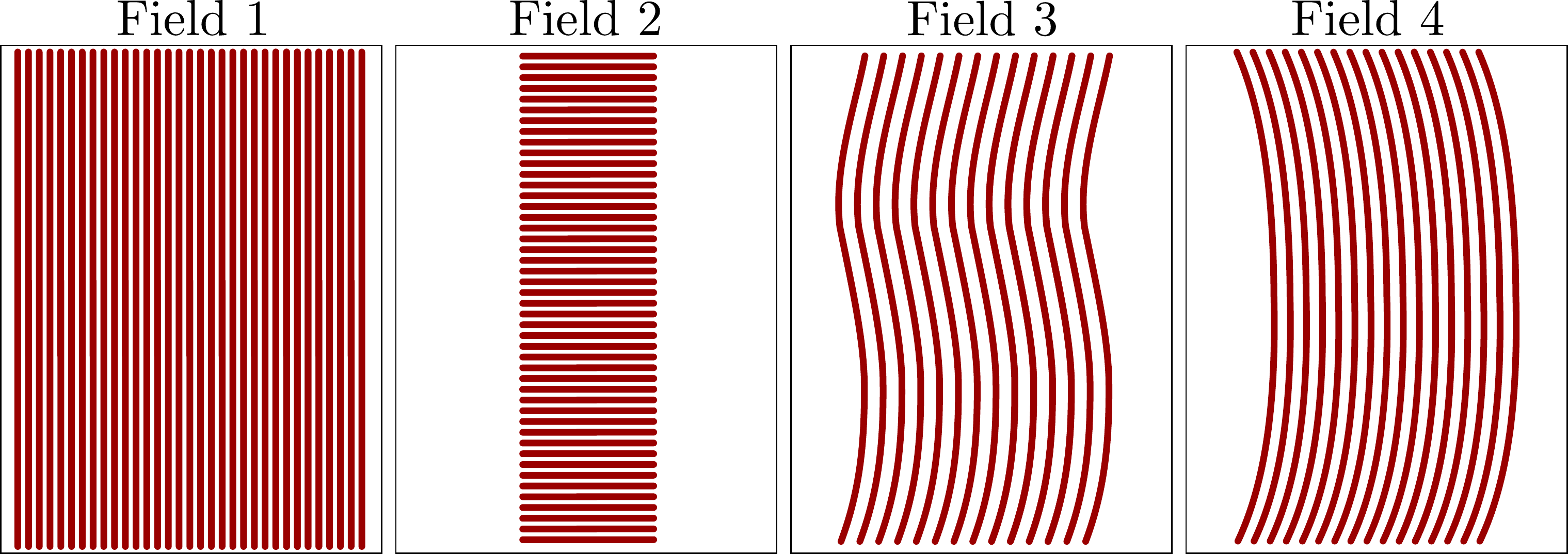}
\caption{Fields with different shapes used in the experiments. 
Field~1 is large and long; 
Field~2 is short and requires many turns in quick succession; 
Field~3 is S-shaped;
Field~4 is parabola shaped.}
\label{fig:fieldshapes}
\end{figure} 

\begin{table}[t]
\centering
\begin{tabular}{@{}cclcc@{}}
\toprule
\multicolumn{1}{c|}{Field 1} & \multicolumn{2}{c}{0.651 $\pm$ 0.826 cm} & 2.28 & 100\% \\
\multicolumn{1}{c|}{Field 2} & \multicolumn{2}{c}{0.872 $\pm$ 1.534 cm} & 2.17 & 100\% \\
\multicolumn{1}{c|}{Field 3} & \multicolumn{2}{c}{0.814 $\pm$ 1.144 cm} & 2.46 & 100\% \\
\multicolumn{1}{c|}{Field 4} & \multicolumn{2}{c}{0.830 $\pm$ 1.155 cm} & 4.38 & 100\% \\ \midrule
\multicolumn{1}{c}{} & \multicolumn{2}{c}{\begin{tabular}[c]{@{}c@{}}Avg. and std.\,dev.\\ distance to crop rows\end{tabular}} & \begin{tabular}[c]{@{}c@{}} Avg. missed\\crops\,per\,row \end{tabular} & \begin{tabular}[c]{@{}c@{}}Percentage\,of\\ visited rows\end{tabular} \\ \bottomrule
\end{tabular}
\caption{Field coverage analysis in the fields illustrated in~\figref{fig:fieldshapes} using our navigation scheme.}
\label{tab:exp2}
\end{table}

The second experiment is designed to evaluate the capability of a robot using
our navigation scheme to cover fields of different shapes and lengths. We
consider four different fields to evaluate our navigation scheme, which are shown
in~\figref{fig:fieldshapes}. Field 1 presents a typical scenario having long
crop rows, whereas Field 2 is a short but wide,  which requires the robot to
turn several times in quick succession. Field 3 and 4 have curved crop rows
which are typically found in real world fields. To evaluate the navigation
performance, we consider the following metrics: (i)~the average and standard
deviation of the distance of the robot from the crop rows, (ii)~the average
number of crops per row missed by the robot and, (iii)~the percentage of crop
rows in the field missed by the robot during navigation.

\tabref{tab:exp2} summarizes the performance of our navigation scheme for each
of the four fields. The robot is able to navigate along all of the fields with
an average distance from the crop rows of 0.8\,cm and a standard deviation of
1.15\,cm (without relying on any map or an external positioning system). This
shows that the robot was able to follow the crop rows closely during
traversal. The number of crops covered by the robot is computed by considering
all of the crops that are within a range of 10\,cm from the trajectory of the
robot. This threshold ensures that the crops we are monitoring are visible in
the center region of the image for our robot setup. In all of the fields, the
average number of plants missed per row is negligible with respect to the
number of plants in a real crop row. These missed crops are the ones which the
robot misses while entering the next row as shown in~\figref{fig:exp1:traj}.

Finally, we also evaluate the number of crop rows in the field that were
visited by the robot. We consider a row to be visited only if the robot misses
less than 5 crops in a row (i.e. it does not take a shortcut at the end of the
row). For each of the four fields, the robot was able to traverse all of the
crop rows successfully. These results indicate that our system is able to
monitor the crops in the field with a high coverage even in fields presenting
different characteristics.

\subsection{Robustness to User-Defined Parameters}

\begin{figure}[t]
\centering
\subfigure{\includegraphics[width=\linewidth]{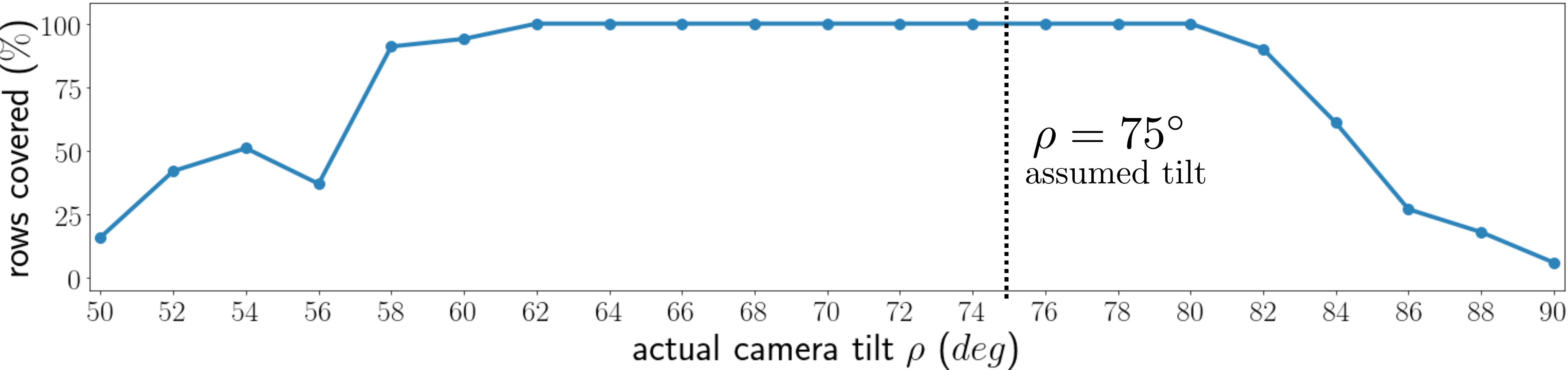}}
\subfigure{\includegraphics[width=\linewidth]{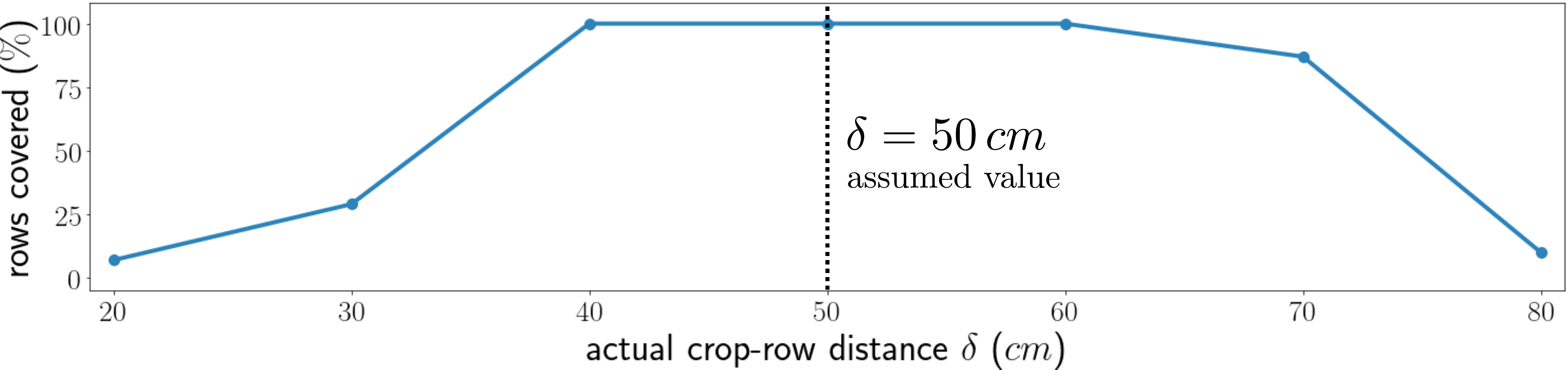}}
\caption{Robustness of our navigation system to errors in camera tilt~$\rho$~(top) and crop-row distance~$\delta$~(bottom) with respect to the assumed values. 
}
\label{fig:robustness}
\end{figure} 

In this experiment, we evaluate the robustness of our navigation scheme to the
critical parameters which needs to be provided by the user. The first
parameter is the camera tilt angle $\rho$ which is used by the visual servoing
controller for following the crop row. Another critical parameter is the crop
row distance $\delta$ which the navigation scheme uses for estimating the
shift of the window~$\mathcal{W}$ at the end of the row. The crop row distance
$\delta$ may not be accurately measured or varies in different parts of the
field. Therefore, to account for these errors, we analyzed the robustness of
our navigation scheme to the camera tilt angle~$\rho$ and the row
distance~$\delta$.

To analyze the robustness of the navigation scheme, we use the percentage of the
crop rows in the field traversed by the robot. This measure is indicative of how
successful the robot is in transitioning to the next row. To test the first
parameter, we fix the camera tilt~$\rho$ in the visual servoing controller
to~75\degree and vary the actual tilt of the camera on the robot in the range
from~50\degree to~90\degree. In \figref{fig:robustness}~(top), we observe that
the robot is able to traverse all the crop-rows in the field (100\% coverage)
for~$\rho$ in the range from~62\degree to~80\degree. This range corresponds to
an error in the camera tilt varying from~-13\degree to~+5\degree with respect to
the assumed value. Thus, we suggest to know the tilt parameter up to $\pm$ 5\degree 
which is easy to achieve in practice.

For the second parameter, we assume the crop row distance~$\delta$ to be
50\,cm in the controller and compute the shift for the window $\mathcal{W}$
based on this value. We evaluate the robustness of the system to this
parameter by considering fields with a crop-row distances~$\delta$ that range
from~20\,cm to 80\,cm. We observe in \figref{fig:robustness}~(bottom) that the
robot is able to perform all the transition to the next rows successfully for
$\delta$ varying from~40\,cm to~60\,cm. This corresponds to a difference
of~-10\,cm to~+10\,cm from the assumed~$\delta$ which is a reasonable
variation considering that most fields are sown with precision agricultural
machines today. These results indicate that our system it is robust
to reasonable error which is expected in real-world scenarios.

\subsection{Demonstration on Real Robot}

\begin{figure}[t] 
\centering 
\fbox{\includegraphics[width=\linewidth]{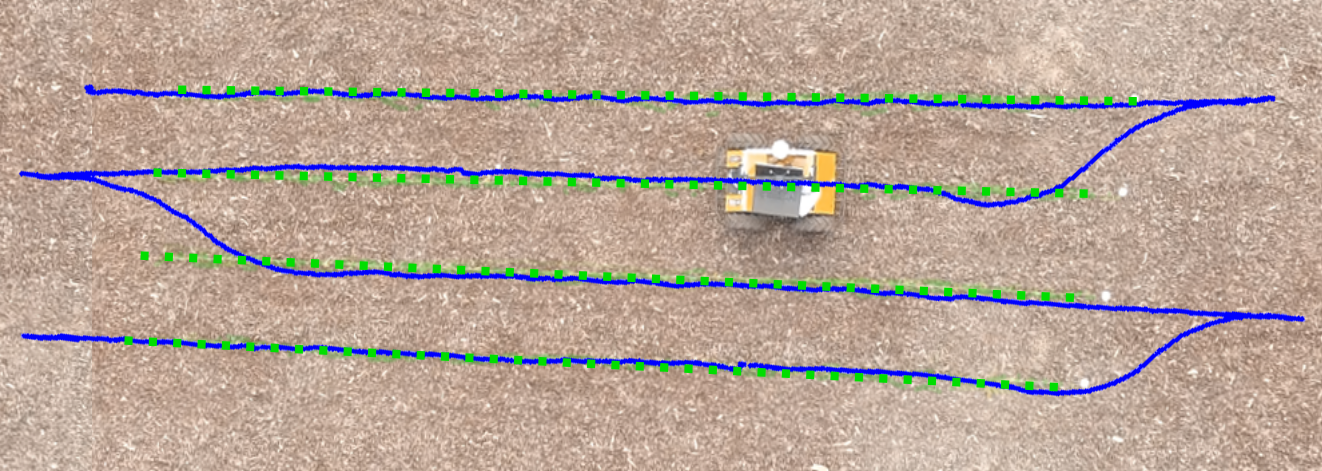}}
\caption{Real robot following our visual-based navigation scheme to navigate
in a test row-crop field, and the trajectory~(blue) of the robot recorded with a RTK-GPS.}
\label{fig:agribot}
\end{figure}

In the last experiment, we demonstrate our navigation system running on a real
robot operating in an outdoor environment.  For this experiment, we used a
\textit{Clearpath Husky} robot equipped with  two  cameras  arranged in the
configuration shown in~\figref{fig:motivation}. All the computations were
performed on a consumer grade laptop running~ROS.  We setup a
test-field with 4 parallel rows each $15$ meters on a rough terrain. The robot
was able to successfully navigate along the crop-rows and switch to the
correctly the end of each crop row. We recorded the robot's trajectory  using
a RTK-GPS system but only to visualize it by overlaying on a aerial image of
the  field as shown in~\figref{fig:agribot}. We observed that the robot
traverses the rows by navigating close to the crops rows within a deviation
of~4\,cms and transition to the next rows within a average maneuvering length
of~1.2\,m at  the start/end of each row.

\section{Conclusion}
\label{sec:conclusion}

In this paper, we presented a novel approach for autonomous robot navigation in crop fields, which
allows an agricultural robot to carry out precision agriculture tasks such as
crop monitoring. Our approach operates only on the local observations from the
on-board cameras for navigation. Our method exploits the row structure
inherent in the crop fields to guide the robot along the crop row without the
need for explicit localization system, GNSS, or a map of the environment. It handles
the switching to new crop rows as an integrated part of the control loop. This
allows the robot to successfully navigate through the crop fields row by row and
cover the whole field. We implemented and evaluated our approach on different
simulated datasets as well as on a self-built agricultural robot. The
experiments suggest that our approach can be used by agricultural robots in
crop fields of different shapes and is fairly robust to the critical user
defined parameters of the controller. 


\clearpage
\bibliographystyle{plain}

\bibliography{glorified,new}

\end{document}